\documentclass{article}
\usepackage[eabstract]{log_2026}

\usepackage{booktabs}
\usepackage{graphicx}
\usepackage[numbers,compress,sort]{natbib}
     \usepackage{wrapfig}
\usepackage{enumitem}
\usepackage{amssymb}
\usepackage{array}
\usepackage[backref=page]{hyperref}
\usepackage[capitalize,noabbrev,nameinlink]{cleveref}

\newcommand{\snapshotdate}{15 October 2026}
\newcommand{\numpapers}{885}
\newcommand{\numdatasets}{111}
\newcommand{\numresults}{27{,}723}

\usepackage{todonotes}

\title[A Living Leaderboard for Graph Machine Learning]{A Living Leaderboard for Graph Machine Learning}

\author[Ritzert]{Martin Ritzert\\Leipzig University, Germany\\\email{martin.ritzert@uni-leipzig.de}}

\begin{document}

\maketitle

\begin{abstract}
    Graph machine learning benchmarking results are scattered across papers that use quite different sets of datasets and baseline methods, making it difficult to tell whether a reported gain reflects architectural progress or merely an outdated comparison or weakly tuned baseline.
    We introduce a living, graph-learning-specific leaderboard that organizes reported results by dataset, putting new results into context.
    As of \snapshotdate{} our LLM-based extraction pipeline extracted \numresults{} eligible results across \numdatasets{} datasets from \numpapers{} peer-reviewed papers. Our analysis of these results quantifies concerns about graph-learning benchmarking: legacy datasets are still actively used and many result tables contain baseline results that are drastically lower than those already available in the literature, e.g. 45\% Amazon-ratings tables and 37\% for Peptides-struct.
The leaderboard allows to quickly compare new results against compatible prior results, aiming to improve the overall quality of graph learning benchmarking. 
    The leaderboard is available at \url{https://graphlearningbenchmarks.github.io}, the code at \url{https://github.com/martinalex000/benchmarking_website}.
\end{abstract}

\section{Why a Living Graph-Learning Leaderboard?}
\label{sec:introduction}

Benchmarking quality in graph machine learning has a long tradition of being questioned, both through large-scale benchmarking efforts \cite{Errica:2020fair,Dwivedi:2023benchmarking,Hu:2020ogb,Katsman:2026mlp} as well as a recent position paper suggesting that weak benchmarks will hinder further progress in graph machine learning~\cite{bechler2025position}. 
Common problems identified in the literature include inadequately tuned baselines \cite{Katsman:2026mlp}, selective choice of baselines to compare against \cite{Errica:2020fair}, as well as the use of outdated datasets \cite{Hu:2020ogb}.
In particular, there is no universally accepted and adopted set of modern graph-learning benchmarks and no tuning process defined, even though the community made progress on this with the OGB~\cite{Hu:2020ogb}, LRGB~\cite{Dwivedi:2022lrgb}, and GraphBench~\cite{stoll2025graphbench} benchmark suites where the latter also includes suggestions on how baselines should be tuned.

We created an LLM-powered pipeline to extract benchmarking results from papers and analyzed \numpapers{} peer-reviewed papers, collected by recursively following incoming and outgoing citation links from a set of seed papers, e.g.\cite{Dwivedi:2022lrgb,Hu:2020ogb}.
Based on the extracted results, we were able to quantify benchmarking problems and trends. In particular, we aim to answer the following questions:
\begin{itemize}[nosep]
    \item \textbf{MLP baselines:} In how many cases did even the best method in a paper not beat a plain MLP?
    \item \textbf{Legacy benchmarks:} How did the use of legacy datasets such as \texttt{ENZYMES} or \texttt{Texas} and \texttt{Squirrel} develop over time? Are they still actively used?
    \item \textbf{Stale baselines:} How often did papers report outdated performance metrics even though better ones were available at publication?
    \item \textbf{Architectural direction:} Did graph transformers take over all the leaderboards?
\end{itemize}

The main contributions of the paper are (i) the living leaderboard itself and (ii) a quantitative analysis of benchmarking problems at scale.
Overall, outdated datasets and stale baselines are a problem at literature scale.
Thus, extended baseline tuning is necessary (see also \cite{bechler2025position,stoll2025graphbench}) and the community should focus on modern datasets.
Through our automated extraction pipeline and planned weekly integration of new papers, our coverage is much higher than manually maintained baselines and allows to track both suggested models as well as baselines and their development over time.

\subsection{Related Work}\label{sec:relatedwork}
Several leaderboards exist throughout machine learning, with the most well-known one being \url{paperswithcode.co} and an LLM-specific example being \url{artificialanalysis.ai/leaderboards/models}.
Within the graph-learning community, OGB maintains a leaderboard for their datasets \cite{Hu:2020ogb}, which is manually maintained and misses a few recent results found by our pipeline.
As for benchmarking problems in graph ML, the position paper \cite{bechler2025position} summarizes major issues. Other relevant work includes benchmarking efforts \cite{Errica:2020fair}, the observation that MLPs are surprisingly strong \cite{Katsman:2026mlp}, and proposals for higher benchmarking standards \cite{Hu:2020ogb,Dwivedi:2023benchmarking,stoll2025graphbench}.

\section{Leaderboard Features and Implementation}
\label{sec:leaderboard}
\begin{figure}[t]
    \includegraphics[width=\textwidth]{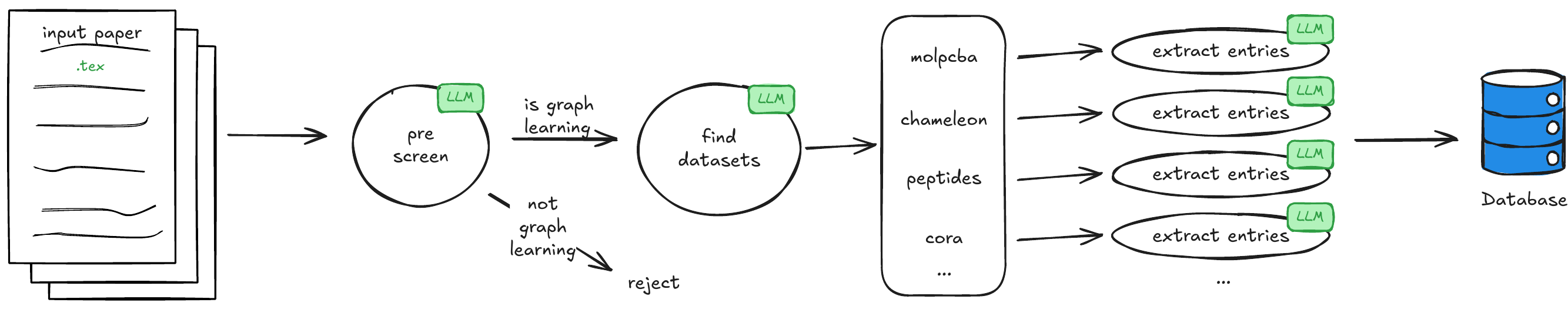}
    \caption{LLM-powered extraction pipeline. All three round operations are based on LLM calls. The first two are single calls, while the final extraction is one call per detected dataset.}
    \label{fig:extraction_pipeline}
\end{figure}
\begin{wrapfigure}{r}{0.4\textwidth}
    \vspace{-.4cm}
    \includegraphics[width=\linewidth]{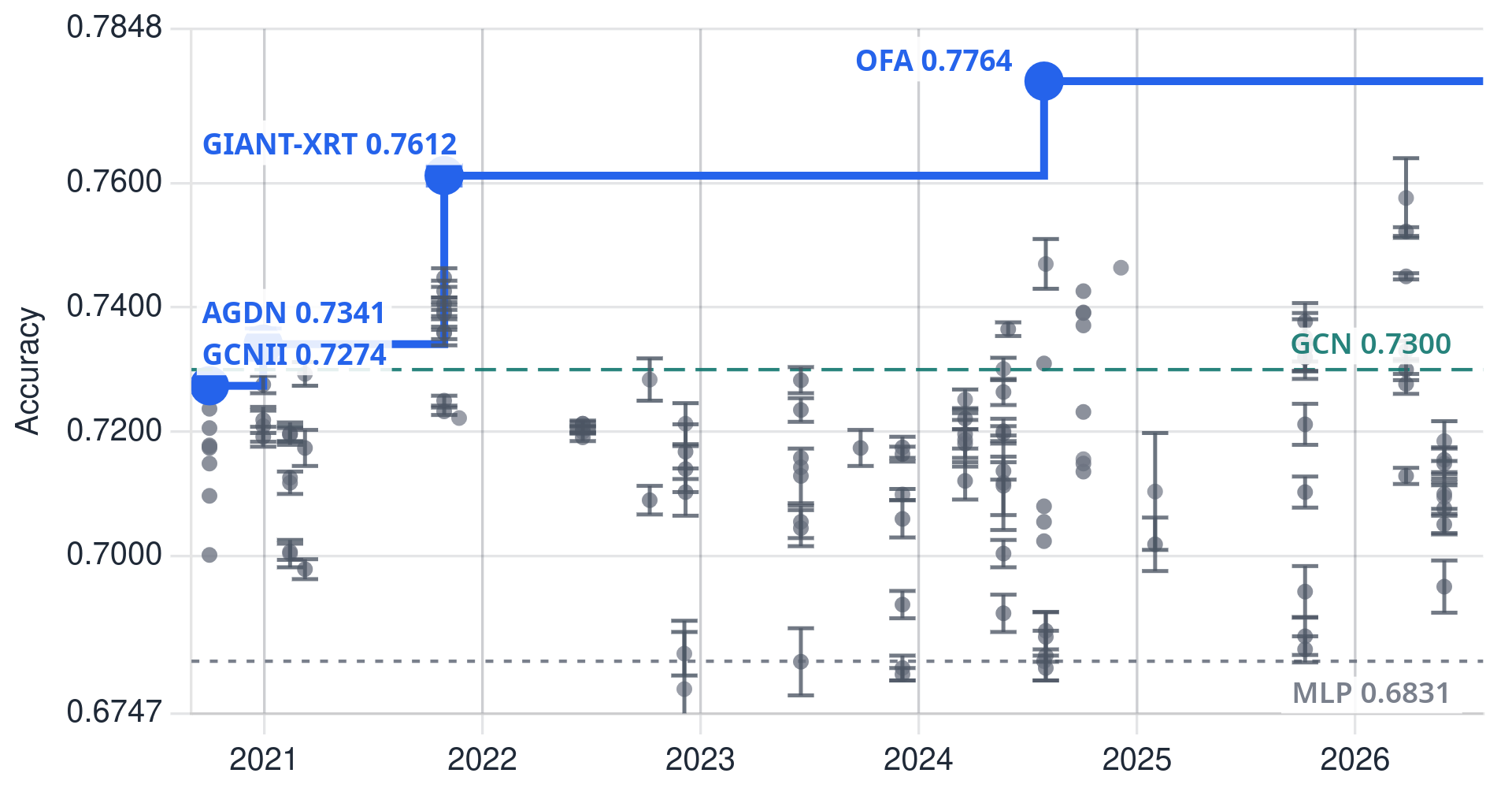}
    \caption{SOTA-over-time plot for \texttt{ogbn-arxiv} highlighting the SOTA, as well as GCN and MLP baselines.}
    \label{fig:sota-over-time}
    \vspace{-0.5cm}
\end{wrapfigure}

Our leaderboard contains two main types of pages: dataset pages that list all benchmarking results for that dataset and papers pages that display the results of the paper in context with GCN and MLP baselines (if missing) and the top 3 leaderboard entries.
The dataset pages contain a plot showing the SOTA over time together with baselines (see \cref{fig:sota-over-time}) and a filterable table with results and their sources. 
Additionally, the often competitive performance of simple MPGNNs and MLPs are highlighted.

\paragraph{LLM-powered extraction.}

Our extraction pipeline proceeds through (i) candidate screening, (ii) dataset discovery, (iii) table selection, and (iv) result extraction as shown in \cref{fig:extraction_pipeline}.
Candidate screening is based on the title and abstract only and asks whether the paper is about graph learning and thus may plausibly contain graph machine learning benchmark results.
After the initial relevance screening, we download the arXiv source (a pdf pipeline is under development) and identify table environments.
We then send the introduction and experiment section, table captions, first row and column of each table into the LLM (we used \texttt{Gemma4-26B-A4B}, 4-bit) to identify which table contains results for which dataset.
Based on this information and the chosen tables, we then extract result entries (including stds while skipping ablations) per dataset and store them in a database.
Rows only enter a leaderboard when the extracted task matches.
Custom, OOD, few-shot, continual, and attack settings are excluded.
We additionally extract metadata such as codebase URLs or the type of each method (MPGNN, GT, MLP) which can be used for filtering.

\section{Analysis}
\label{sec:analysis}

All analyses use a snapshot from \snapshotdate{}.
On that date, the curated peer-reviewed corpus contained \numpapers{} papers, \numdatasets{} datasets, and \numresults{} eligible extracted results.
When both results report a standard deviation, we use a two-sided Welch's t-test at the 5\% significance level. When only one result reports a standard deviation, we compare the other value, treated as fixed, with the 95\% confidence interval for the mean. When neither result reports a standard deviation, unequal values are treated as significantly different. We do not correct for multiple comparisons. When repetition counts are unavailable, we assume four runs, which we assume to be a minimum bar for stds to be reported in the first place. This assumption is conservative when the actual (unavailable) count is larger, as it generally makes differences harder to become significant.

\subsection{The Importance of Node Features (MLP Performance)}
\label{sec:mlp}

\begin{figure}[t]
    \centering
    \includegraphics[width=\linewidth]{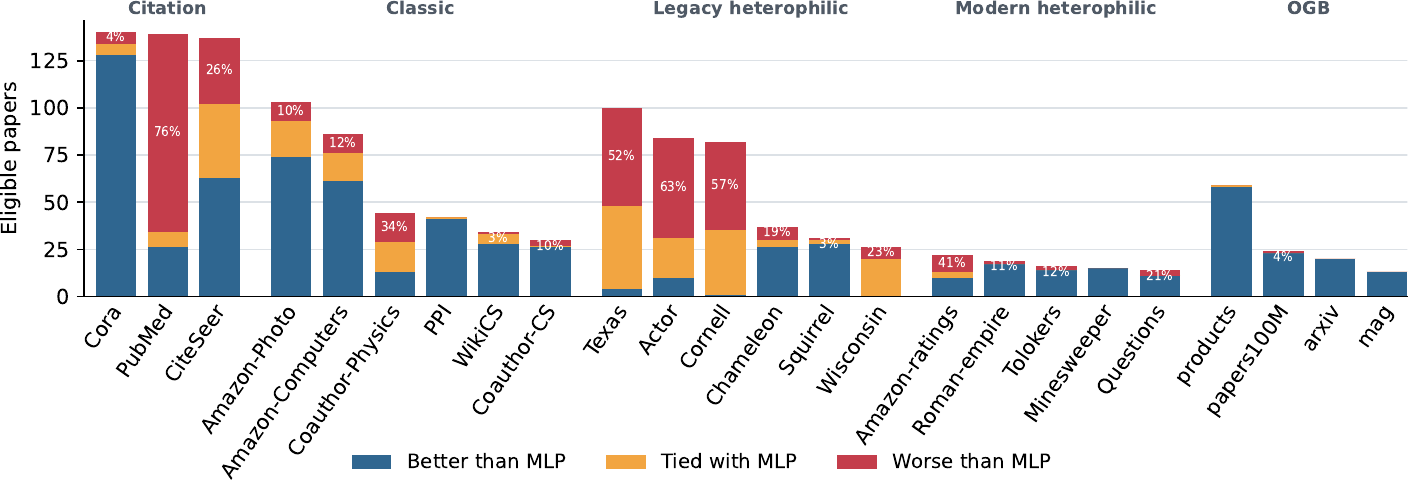} 
    \caption{Best non-MLP method per paper against the overall best MLP (as of today). Especially on text-heavy legacy datasets often no reported result in a paper outperforms a plain MLP.}
    \label{fig:mlp}
\end{figure}

MLPs provide a surprisingly hard-to-beat baseline on a large number of node classification datasets \cite{Katsman:2026mlp,zhang2021graph}.
We thus analyzed for each dataset the fraction of papers where non-MLP methods are better, tied, or worse than the best eligible MLP result in the literature (irrespective of publication date, as MLPs are an established baseline).
In \cref{fig:mlp} we see that especially on the legacy heterophilic datasets \texttt{Texas}, \texttt{Cornell}, \texttt{Actor}, and \texttt{Wisconsin}, graph learning techniques are not providing any benefits over plain MLPs.
This also holds for other datasets with textual node features such as \texttt{PubMed} or \texttt{Coauthor-Physics/CS} (even without LLM embeddings).
One should thus first take MLPs seriously as baselines in graph machine learning and second, as pointed out by \mbox{\citet{Katsman:2026mlp}} and avoid using datasets where the graph structure is not important to track progress in graph ML.

\subsection{Usage of Legacy Datasets}
\label{sec:datasets_over_time}

\begin{wrapfigure}{r}{0.43\textwidth}
    \centering
    \includegraphics[width=\linewidth]{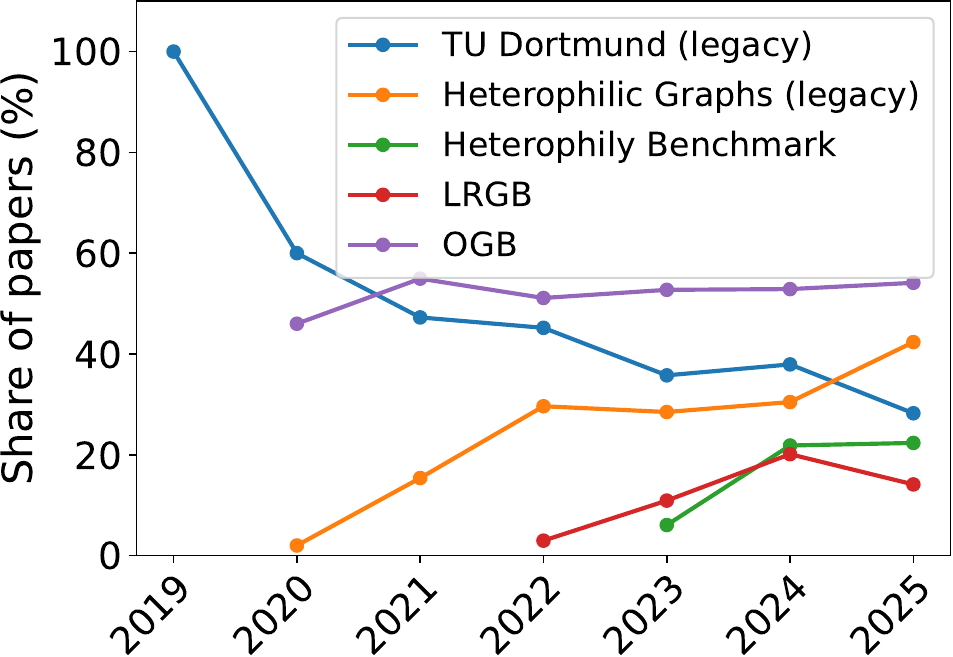}
    \caption{Dataset usage over time. Legacy datasets are still actively used.}
    \label{fig:datasetsOverTime}
\end{wrapfigure}
Over the last years, several papers have introduced new and modern datasets of high quality, such as OGB \cite{Hu:2020ogb} or recently GraphBench \cite{stoll2025graphbench}.
We thus ask to what extent dataset usage has been shifting from TU Dortmund datasets \cite{morris2020tudataset} to those newer ones.
The (legacy) heterophilic datasets \texttt{Texas}, \texttt{Squirrel}, etc. \cite{pei2020geom} contain problems (size, test data leakage, MLP performance), and \texttt{Roman-Empire}, \texttt{Tolokers}, etc. should be used instead \cite{platonov2023critical}.
We are thus interested in how far usage has shifted from the two legacy datasets to modern alternatives and plot the number of papers reporting results on at least one of those in \cref{fig:datasetsOverTime} (papers may contribute to multiple datasets).
On our collected data we observe a diverging picture: while TU datasets are being used less, legacy heterophilic graphs are now used by 42.4\% of 2025 conference papers using at least one of the five plotted benchmark groups, making them the second-most-used group.
As those datasets are ill-suited to measure graph-specific progress \cite{Katsman:2026mlp,platonov2023critical} this is an alarming trend.

\subsection{Baseline Gaps}
\label{sec:stale_baselines}

\begin{figure}
    \centering
    \includegraphics[width=\linewidth]{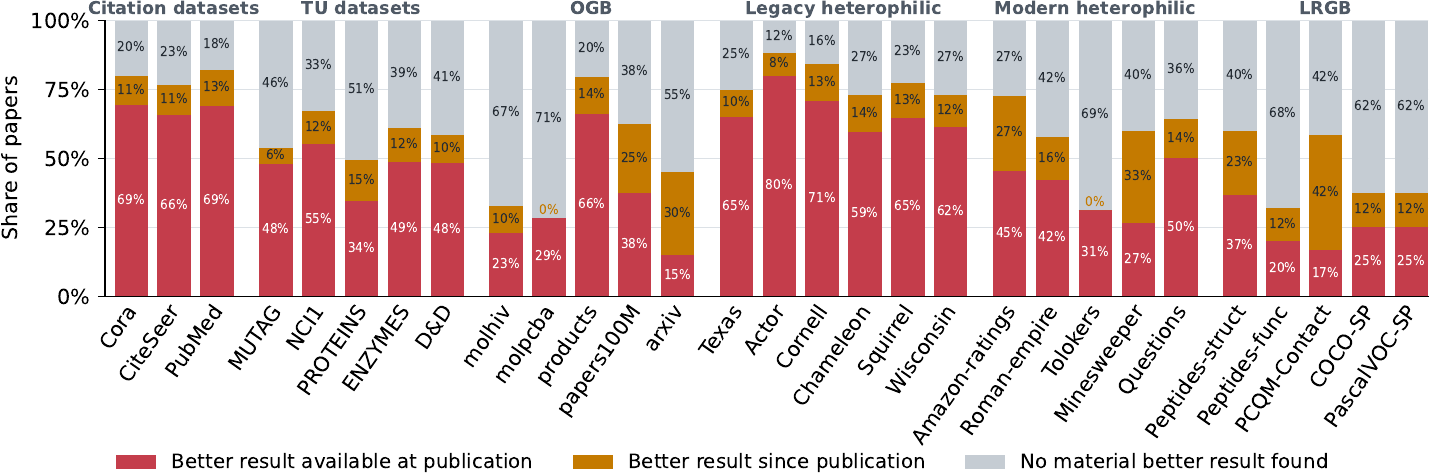} 
    \caption{Fraction of papers where at least one baseline had better results elsewhere. 
}
    \label{fig:stale_baselines}
\end{figure}

Gaps between reported results and corresponding literature values can inflate apparent architectural progress.
(e.g. due to suboptimal hyperparameter configurations)
An extreme example was shown by \mbox{\citet{bechler2025position}} where the SOTA of \texttt{ENZYMES} dropped according to several papers.
Quantifying this, in \cref{fig:stale_baselines} we count in how many papers a baseline is reported weaker than its corpus value at time of publication (red) and overall (orange). In our collected corpus, 67.1\% of analyzed papers include at least one result with a significantly better value already recorded for the same method in another paper at publication.
The observed rates are lower for OGB, which has a manually maintained leaderboard and thus an easy way of looking up the SOTA, and for the relatively recent LRGB benchmark.
\begin{wrapfigure}{r}{0.38\linewidth}
    \centering
    \includegraphics[width=\linewidth]{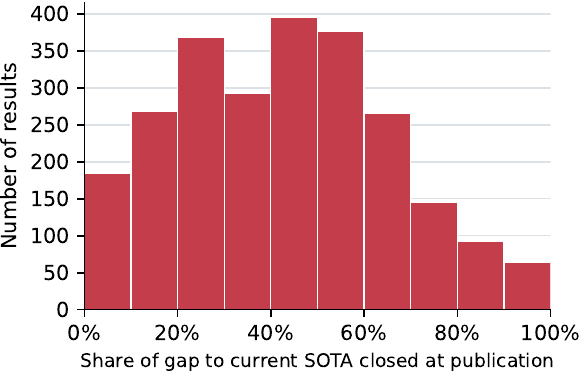}
    \caption{Relative reduction in a method’s distance to the SOTA using its corpus value instead of its paper value}
    \vspace{-.5cm}
    \label{fig:gap_closed}
\end{wrapfigure}
Generally, the high values observed in our corpus indicate a general problem in benchmarking.
Appendix \cref{fig:stale_baselines_appendix} shows the same for individual results and reaching up to 29.6\% results lower than their corpus value (on \texttt{Actor}), underscoring the need for both high rigor while tuning baselines as well as a centralized place to look up benchmarking results. 
To measure severity, we compute for each method \(x\) the relative reduction in its distance to the current SOTA when replacing the paper-reported value with its corpus value: \(\frac{|\text{Paper}(x)-\text{Corpus}(x)|}{|\text{Paper}(x)-\text{SOTA}|}\), where SOTA is the best recorded performance on the dataset across all methods.
The histogram in \cref{fig:gap_closed} shows that in 38.5\% of cases the best literature value was at least halfway to the SOTA, i.e. the (baseline) method is a lot better than the paper's table suggested.

\subsection{Message Passing vs. Graph Transformers}
\label{sec:graph_transformers}

\begin{wrapfigure}{r}{0.46\textwidth}
    \centering
    \includegraphics[width=\linewidth]{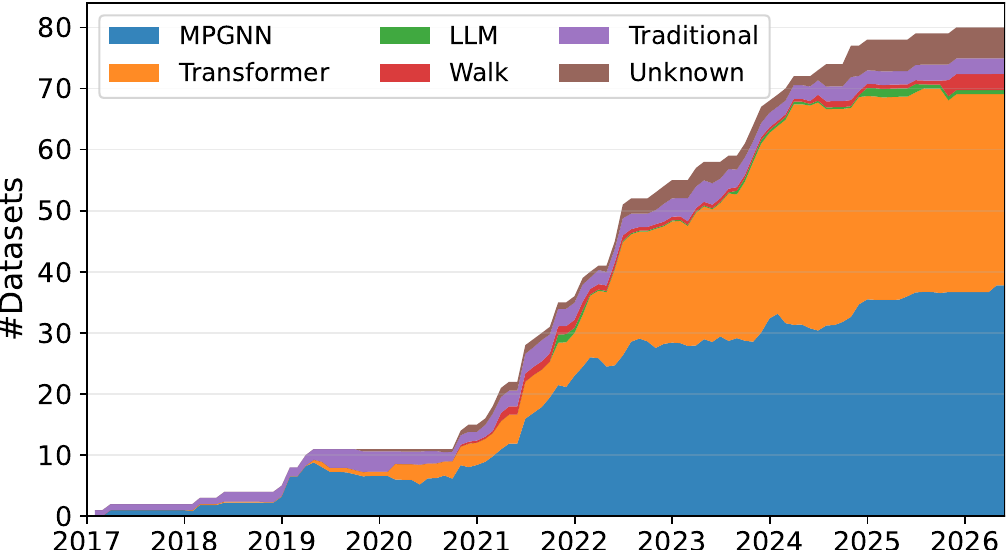}
    \caption{Architecture of best models over datasets. MPGNNs are leading but not dominating.}
    \label{fig:gnn_vs_gt}
\end{wrapfigure}
There has been a long-running debate over whether graph transformers would take over graph-learning leaderboards, as transformers did in other areas.
We thus count how many datasets (with a minimum of five papers reporting on them) are led by each architecture category, e.g. message-passing GNNs, graph transformers, LLMs, or traditional methods such as SVMs and boosting.
Each dataset has unit weight, distributed evenly among all methods treated as tied for being the best.
With 39.1\% GTs vs 47.2\% MPGNNs, neither category dominates and the picture remained stable for the last two years, see \cref{fig:gnn_vs_gt}.
MPGNNs lead on node-level tasks, while GTs have a slightly larger share on graph-level tasks (see Appendix).
We can also observe that traditional methods are still strong contenders on several datasets, especially around molecules.

\section{Limitations, Future Directions, and Conclusion}
\label{sec:limitations}

Around 80\% of NeurIPS, ICLR, and ICML papers from 2022--2024 are available on arXiv (\cref{tab:arxiv-share} in the appendix), limiting source-based coverage.
We thus plan to develop a PDF extraction path for papers without available \LaTeX{} sources.
The second limitation is that incompatible results may be merged into the same leaderboard, e.g. from different evaluation protocols or different metrics.
Figuring out whether a result is really compatible is non-trivial even for humans, and exactly where we plan to improve the current pipeline.
Until then, we have guards in place (same metric, outlier detection, manual exclusion, rules against typical cases) avoiding most incompatible results.

\label{sec:conclusion}

In this paper, we presented a leaderboard specifically for graph machine learning that allows users to quickly identify the best baselines for any given dataset.
Within our collected corpus, we found that legacy datasets are even gaining popularity and that there are often better results for reported baselines, especially for MLPs: for example, on 57.3\% of papers reporting \texttt{Cornell} results, the best known MLP outperforms all reported non-MLP methods in that paper on that dataset.
Further, neither GTs nor MPGNNs dominate the observed frontier.
Overall, we observed rising use of legacy datasets with known problems, stronger-than-expected MLP baselines, and regular reporting of results for which the corpus contains a better one under the same name.

\bibliographystyle{unsrtnat}
\bibliography{reference}

\appendix

\section{Technical Details}

The extraction is based on LLMs and in our case a 4-bit-quantized \textsc{Gemma-4 26B-A4B-it} instance was selected as a fast and cost-effective model for structured extraction.
We also tested against smaller and larger families (gpt-oss, Qwen, GLM, MiniMax, see \cref{tab:extraction-example}) and found that Gemma and GLM both generally worked well, but prompts have to be adapted to the model at hand to reliably ignore ablation tables and properly normalize method names.
For the initial abstract screening, we relied on a \textsc{gpt-oss-120b} model, which is considerably faster than our Gemma instance.
In addition to the extraction pipeline, we tried a simple one-shot approach, essentially telling the model to extract all benchmark results, which turned out to be less reliable.
For extracting the number of repetitions for each dataset, a more advanced model was necessary where we used \textsc{GLM-5.3-flash}.

In order to render the website, we export the results to Markdown and then use Hugo to generate a static website.
The SOTA-over-time plot as well as the filtering and sorting of tables are done client-side using JavaScript.

We ran our experiments on a server with EPYC 7352 CPUs and 1\,TB RAM.
In general, the hardware requirements (except for running the LLM) are quite low and it would have been possible to run the experiment on a laptop with API access to a language model.
Our language models were run using vLLM on two NVIDIA A100 40\,GB cards.
The current extraction pipeline is relatively inefficient with respect to token usage; a full re-evaluation takes approximately 3 days at $\sim$20 papers/h ($\sim$70 papers evaluated in parallel to boost throughput).

\section{Selection of Papers to Be Parsed}
\begin{wraptable}{r}{0.39\textwidth}
\caption{Share of Semantic Scholar-indexed conference papers with an arXiv identifier.}
\label{tab:arxiv-share}
\centering
\small
\begin{tabular*}{\linewidth}{@{\extracolsep{\fill}}lccc@{}}
\toprule
\textbf{Conference} & \textbf{2022} & \textbf{2023} & \textbf{2024}  \\
\midrule
NeurIPS & 79.5\% & 79.4\% & 80.3\% \\
ICLR    & 80.4\% & 83.1\% & 84.2\% \\
ICML    & 83.0\% & 84.2\% & 80.2\% \\
\bottomrule
\end{tabular*}
\end{wraptable} 
Our corpus consists of conference papers that are on arXiv since the available \LaTeX source makes parsing tables easier.
We plan to extend the pipeline to include pdfs directly through e.g. OpenReview to get complete coverage of major ML conferences.
About 80\% of Semantic Scholar-indexed NeurIPS, ICLR, and ICML papers from 2022--2024 have an arXiv identifier, see \cref{tab:arxiv-share}. 
To build the corpus, we relied on citation links and Semantic Scholar, initialized with graph-learning papers including LRGB~\cite{Dwivedi:2022lrgb} and GraphBench~\cite{stoll2025graphbench}.
The corpus was then expanded through the papers' citation neighborhoods, following both references and citations whenever an arXiv version was available (\texttt{externalIds.ArXiv}).
A candidate was excluded only if its title and abstract did not pass the graph-learning relevance pre-screen, or if full-text processing yielded no numeric benchmark result compatible with a canonical dataset--task pairing tracked by the leaderboard.

\section{Validation of the LLM pipeline}

Since the pipeline is fully automatic, we tested its performance on multiple ways. 
This included a simple check that modern LLMs are indeed capable of extracting numbers from tables (\cref{sec:extractingNumbers}), a check of the pre-screening method (\cref{sec:preScreeningEval}), as well as an evaluation of the extraction pipeline covering datasets and methods in \cref{sec:datasetExtractionEval}.
Further, there is a module that finds the number of repetitions for each experiment on a table level. Together with the information on whether a result has been trained by the paper themselves, we use this as the basis for our statistical evaluation.

\subsection{Extraction Agreement Case Study}
\label{sec:extractingNumbers}
\begin{table*}[t]
\centering
\small
\setlength{\tabcolsep}{3.5pt}
\caption{Example extracted results on Penn94 from \cite{lim2021large}. While LLMs disagree on which methods to extract (skipping variants, but synchronized), they show perfect reconstruction of numbers.}
\label{tab:extraction-example}
\resizebox{\textwidth}{!}{\begin{tabular}{l>{\itshape}cccccccc}
\toprule
Method & \itshape Human & GLM-5.2 & Gemma-4-26B-A4B & Gemma-4-31B & GPT-OSS-120B & MiniMax-M3 & Llama-3.3-70B & Values match \\
\midrule
APPNP & 0.7433 & 0.7433 & 0.7433 & 0.7433 & 0.7433 & 0.7433 & 0.7433 & \textcolor{green!50!black}{\checkmark} \\
C\&S 1-hop & 0.7428 & 0.7428 & 0.7428 & 0.7428 & 0.7428 & 0.7428 & 0.7428 & \textcolor{green!50!black}{\checkmark} \\
C\&S 2-hop & 0.7840 & 0.7840 & 0.7840 & 0.7840 & 0.7840 & 0.7840 & 0.7840 & \textcolor{green!50!black}{\checkmark} \\
GAT & 0.8153 & 0.8153 & 0.8153 & 0.8153 & 0.8153 & 0.8153 & 0.8153 & \textcolor{green!50!black}{\checkmark} \\
GATJK & 0.8069 & 0.8069 & 0.8069 & 0.8069 & 0.8069 & 0.8069 & 0.8069 & \textcolor{green!50!black}{\checkmark} \\
GCN & 0.8247 & 0.8247 & 0.8247 & 0.8247 & 0.8247 & 0.8247 & 0.8247 & \textcolor{green!50!black}{\checkmark} \\
GCNII & 0.8292 & 0.8292 & 0.8292 & 0.8292 & 0.8292 & 0.8292 & 0.8292 & \textcolor{green!50!black}{\checkmark} \\
GCNJK & 0.8163 & 0.8163 & 0.8163 & 0.8163 & 0.8163 & 0.8163 & 0.8163 & \textcolor{green!50!black}{\checkmark} \\
GCNJK-Cluster & 0.6999 & 0.6999 & 0.6999 & --- & 0.6999 & --- & --- & \textcolor{green!50!black}{\checkmark} \\
GCNJK-SAINT-Node & 0.7280 & 0.7280 & 0.7280 & --- & 0.7280 & --- & --- & \textcolor{green!50!black}{\checkmark} \\
GCNJK-SAINT-RW & 0.7229 & 0.7229 & 0.7229 & --- & 0.7229 & --- & --- & \textcolor{green!50!black}{\checkmark} \\
GPR-GNN & 0.8138 & 0.8138 & 0.8138 & 0.8138 & 0.8138 & 0.8138 & 0.8138 & \textcolor{green!50!black}{\checkmark} \\
L Prop 1-hop & 0.6321 & 0.6321 & 0.6321 & 0.6321 & 0.6321 & 0.6321 & 0.6321 & \textcolor{green!50!black}{\checkmark} \\
L Prop 2-hop & 0.7413 & 0.7413 & 0.7413 & 0.7413 & 0.7413 & 0.7413 & 0.7413 & \textcolor{green!50!black}{\checkmark} \\
LINK & 0.8079 & 0.8079 & 0.8079 & 0.8079 & 0.8079 & 0.8079 & 0.8079 & \textcolor{green!50!black}{\checkmark} \\
LINK Minibatch & 0.8161 & 0.8161 & 0.8161 & --- & 0.8161 & --- & --- & \textcolor{green!50!black}{\checkmark} \\
LINKX & 0.8471 & 0.8471 & 0.8471 & 0.8471 & 0.8471 & 0.8471 & 0.8471 & \textcolor{green!50!black}{\checkmark} \\
LINKX Minibatch & 0.8450 & 0.8450 & 0.8450 & --- & 0.8450 & --- & --- & \textcolor{green!50!black}{\checkmark} \\
MixHop & 0.8347 & 0.8347 & 0.8347 & 0.8347 & 0.8347 & 0.8347 & 0.8347 & \textcolor{green!50!black}{\checkmark} \\
MixHop-Cluster & 0.7579 & 0.7579 & 0.7579 & --- & 0.7579 & --- & --- & \textcolor{green!50!black}{\checkmark} \\
MixHop-SAINT-Node & 0.7561 & 0.7561 & 0.7561 & --- & 0.7561 & --- & --- & \textcolor{green!50!black}{\checkmark} \\
MixHop-SAINT-RW & 0.7638 & 0.7638 & 0.7638 & --- & 0.7638 & --- & --- & \textcolor{green!50!black}{\checkmark} \\
MLP & 0.7361 & 0.7361 & 0.7361 & 0.7361 & 0.7361 & 0.7361 & 0.7361 & \textcolor{green!50!black}{\checkmark} \\
MLP Minibatch & 0.7424 & 0.7424 & 0.7424 & --- & 0.7424 & --- & --- & \textcolor{green!50!black}{\checkmark} \\
SGC 1-hop & 0.6679 & 0.6679 & 0.6679 & 0.6679 & 0.6679 & 0.6679 & 0.6679 & \textcolor{green!50!black}{\checkmark} \\
SGC 2-hop & 0.7609 & 0.7609 & 0.7609 & 0.7609 & 0.7609 & 0.7609 & 0.7609 & \textcolor{green!50!black}{\checkmark} \\
\bottomrule
\end{tabular}}
\end{table*} We found that transcription stability is not an issue with 2026 models (in contrast to early LLMs), and all tested models are capable of successfully copying numbers from a table into JSON.
We illustrate this in \cref{tab:extraction-example}, which shows that different models generally agree on the numerical values and differ mainly in which rows are extracted or skipped and in how method names are normalized.
Through manual verification, we were able to check that indeed all of the values are in the paper and that several models (GLM, Gamma4 MoE, gpt-oss) did extract all existing rows.
In the concrete example, the Minibatching variants ignored by the other models seem to be part of the method and should thus be extracted, but they can easily be mistaken for an ablation and skipped, as those models did.

For more general numbers, we ran the six LLMs \textsc{Gemma-4-26B-A4B} 4-bit (our main driver for the extraction pipeline) as well as \textsc{GLM-5.2}, \textsc{Gemma-4-31B}, \textsc{GPT-OSS-120B}, \textsc{MiniMax-M3}, and \textsc{Llama-3.3-70B} on 10 papers selected to contain relatively many and well-distributed results (2110.14446, 2205.12454, 2305.08018, 2305.13315, 2305.18407, 2306.16976, 2405.02795, 2405.16435, 2502.09263, 2506.04765).
\Cref{tab:extraction-example} is representative of the strong agreement on numerical values observed across the ten-paper evaluation. Outside QM9, only 30 of 16{,}298 pairwise comparisons differed beyond rounding, involving a small number of differences in experimental settings, method variants, and row or column assignments. QM9 initially also exposed unit-normalization issues. The main challenge is consistently deciding which rows represent models to include and which represent variants to exclude, together with normalizing model names, rather than numerical transcription.

\subsection{Pre-Screening}
\label{sec:preScreeningEval}

As detailed in the main paper, the pre-screening is tasked to distinguish graph learning papers from other papers.
For that we first filter based on keywords such as ``graph neural network'' and then send title and abstract of the paper to a fast LLM, in our case \textsc{GPT-OSS-120B}.
We manually evaluated the success of this process by looking (post-hoc) at three different cases: 10 papers accepted by pre-screening that contribute no results, 10 rejected papers that contain specific keywords such as MPGNN or graph embedding, and 10 rejected papers that contain the word graph but none of the specific keywords in title or abstract.
We found that in all 30 cases, manual verification agreed with the LLM screening result: the accepted papers were all about graph learning, but considered tasks that are not yet tracked such as graph generation and adversarial attacks. 
On the other hand, the rejected papers were indeed mostly math papers. 
It turned out that the term ``graph embedding" which we used as one of the filters is regularly used outside of graph ML and was a major driver of papers that made it to the LLM and were rejected afterwards.

While the pre-screening works as expected, only about half of the papers accepted by pre-screening contribute results.
This is because the pre-screening check only very generally checks whether the paper has anything to do with graph learning including areas that are not tracked yet, e.g. graph generation.
Tightening the pre-screening objective could thus make extraction more efficient.

\subsection{Result extraction}
\label{sec:datasetExtractionEval}

This is the main experiment to validate the pipeline.
It is split into two parts: dataset detection and model detection.
For both experiments we evaluated a shared set of 10 papers with results such that we can also estimate the end-to-end precision and recall.

\textbf{Dataset detection:} In our experiment on 10 papers with results we observed 100\% precision and only 62.1\% recall which is mainly due to an overly-pessimistic evaluation of result compatibility since raw YAML extraction achieved a recall of 96.1\% but lower precision.

\textbf{Method detection:}
On the same 10 papers as before, our pipeline was able to detect almost all methods (98.7\% precision, 86.2\% recall). 
The main problems were connected to multiple variants of a method being reported where the extraction pipeline normalized some but not others from the same table. 
For each candidate result on a given dataset, the extraction prompt supplied the canonical task, metric, and split information for that dataset and asked the model to extract this information for each result. Parser-side checks then rejected explicit mismatches.

While the pipeline is tuned to err on the safe side and in doubt reject results, we aim to improve the recall while keeping the precision. 
In our current implementation it is possible (and actually happens) that individual results from a table are marked as incompatible while the remainder of the table makes it into the database.
We plan to remedy that by including an additional verifier step and shifting the compatibility decisions from the result to the dataset level.

\textbf{End-to-End:}
On the same manually annotated 10-paper evaluation, after removing the known incompatible rows, the extraction achieved an end-to-end precision of 99.2\% and 63.9\% recall.
Again, our main focus lies on precision, but it is clear that the recall leaves a lot of room for improvement which we plan to address in the near future.

\textbf{Repetition Counts:}

For the statistical evaluation, we needed the repetition counts behind each reported standard deviation.
For that we ran another extractor, this time based on \textsc{GLM-5.3-flash}, extracting the information on a table-level.
Again, we used 10 papers for manual verification (a different set from above) and found essentially perfect performance.
We then matched the information to the results using another LLM call which we verified on yet another 10-paper set, this time two cross-validation values were not properly propagated with the model erring on the safe side.
By integrating the count extraction directly in the extraction process, the whole matching step is no longer necessary, simplifying our setup.
We note that this will involve switching to a more capable model as the success rate of smaller models on the counts, which are often hidden somewhere in the appendix with highly variable formulations, was not satisfactory.

\section{Additional Results}

\subsection{Message-Passing vs. Graph Transformers}

\begin{figure}
    \centering
    \includegraphics[width=\linewidth]{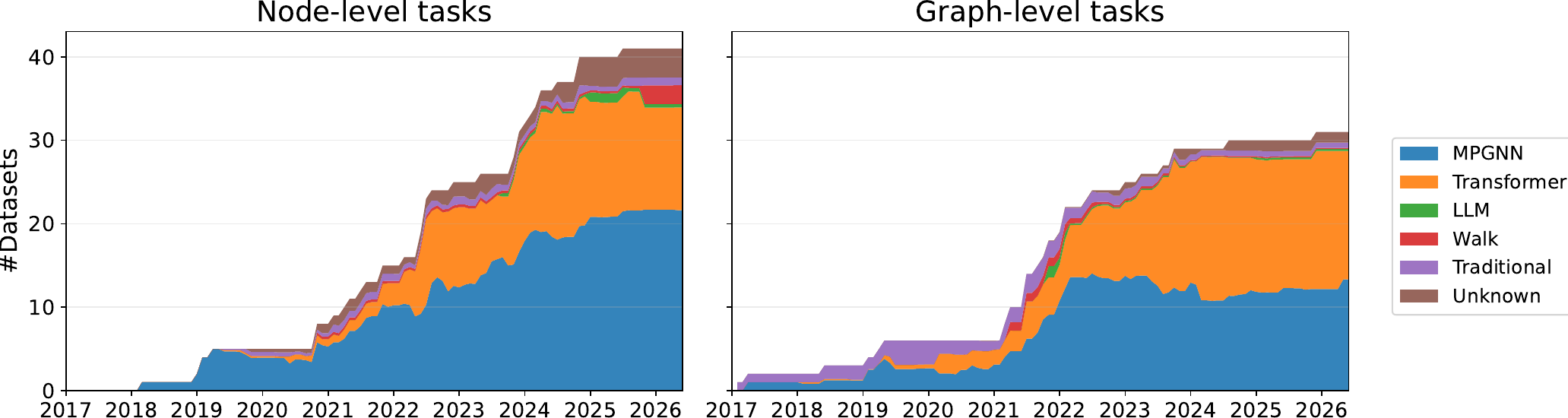}
    \caption{Architecture of best models over graph learning datasets. MPGNNs lead on node-level tasks, while GTs are leading but not dominating on graph-level tasks.}
    \label{fig:gnn_vs_gt_appendix}
\end{figure}

In the main paper, we showed that in general GTs and MPGNNs together dominate graph learning leaderboards with MPGNNs having the larger aggregate share.
With graph learning being split in a multitude of tasks and GTs limited to moderately-sized graphs due to the quadratic scaling of attention, we were interested in how that distributes between node-level and graph-level tasks, shown in \cref{fig:gnn_vs_gt_appendix}.
We observe that MPGNNs lead on node-level tasks (52.8\% versus 30.0\% for GTs), while GTs have a larger share on graph-level tasks (49.6\% versus 43.0\% for MPGNNs).
We thus conclude that neither method is clearly superior to the other and it really depends on the dataset at hand, not only the task.

Note that methods are ``unknown'' if our extraction process was unable to detect the method's type and ``traditional'' includes SVMs, random forests, and all kinds of boosting methods.
For this experiment we considered both pure transformers as well as hybrid architectures such as GPS \cite{rampavsek2022recipe} to be part of the ``transformer'' class while the leaderboard website distinguishes between them.

\subsection{Reported Results vs. Literature Values}

In general, we assumed for this analysis that methods that are reported with the same name are compatible. 
We note that this is not perfectly well-defined, e.g. is a GCN allowed to use skip connections? Or positional encodings? Or virtual nodes? 
For our analysis we generally trust the authors to name methods appropriately and use a different name for bigger modifications of the base method.
We additionally matched methods by their base-name, mapping GCN, GCN$^\star$, and GCN (tuned) to the same ``GCN'' method. 
In general, the method-name normalization is performed by the LLM. 
There, reimplementations by the paper authors and configuration variants representing the same underlying method are consolidated, while qualifiers identifying genuinely distinct methods are preserved.

\begin{figure}
    \centering
    \includegraphics[width=\linewidth]{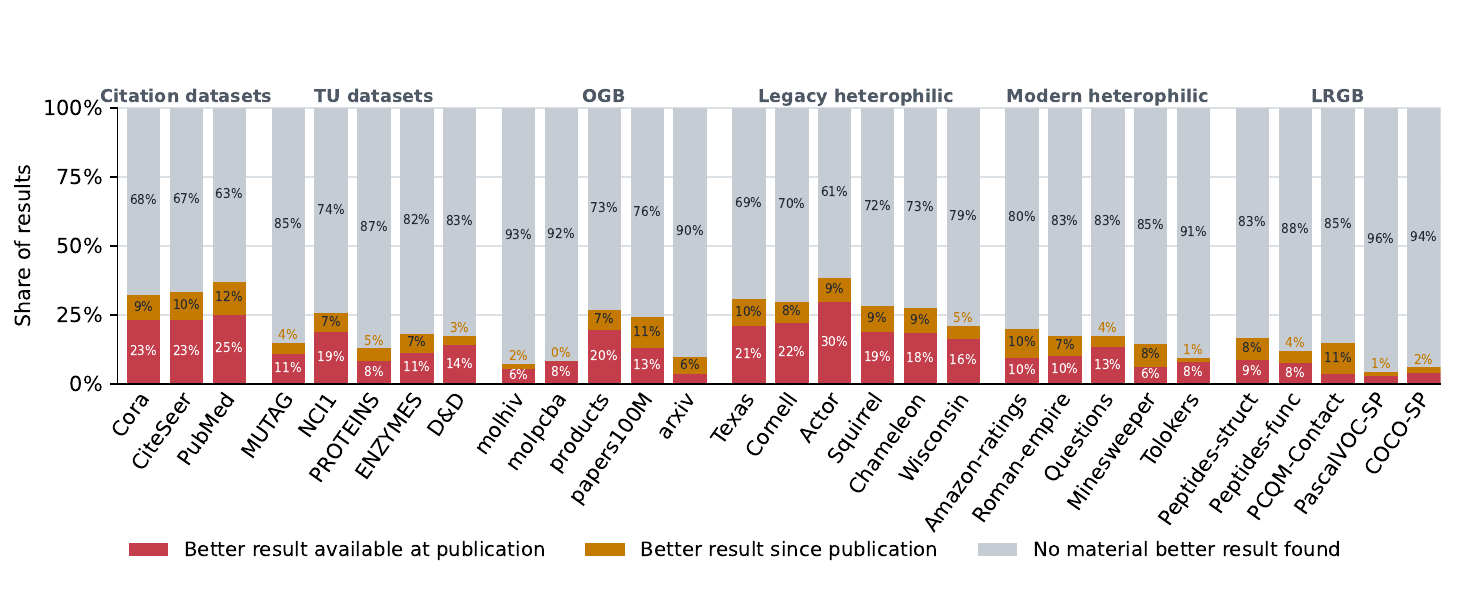}
    \caption{Result comparisons per dataset. Red shows results where a better result was known at publication; orange lists results for which a better result (typically better hyperparameters) has been reported under the same name later.}
    \label{fig:stale_baselines_appendix}
\end{figure}

Here we extend \cref{fig:stale_baselines} by showing individual results instead of aggregating to the paper level: for each dataset, a result is marked in red when a significantly better value for the same method was available at publication, and in orange when a significantly better value was found only since publication.
In \cref{fig:stale_baselines_appendix} we see that it is extremely common to have at least one result which does not represent the literature value.
Across all comparable results, 16.8\% have a significantly better prior literature value, rising to 24.8\% when later literature values are included, clearly showing that the problem is widespread.

\begin{figure}
    \centering
    \includegraphics[width=\linewidth]{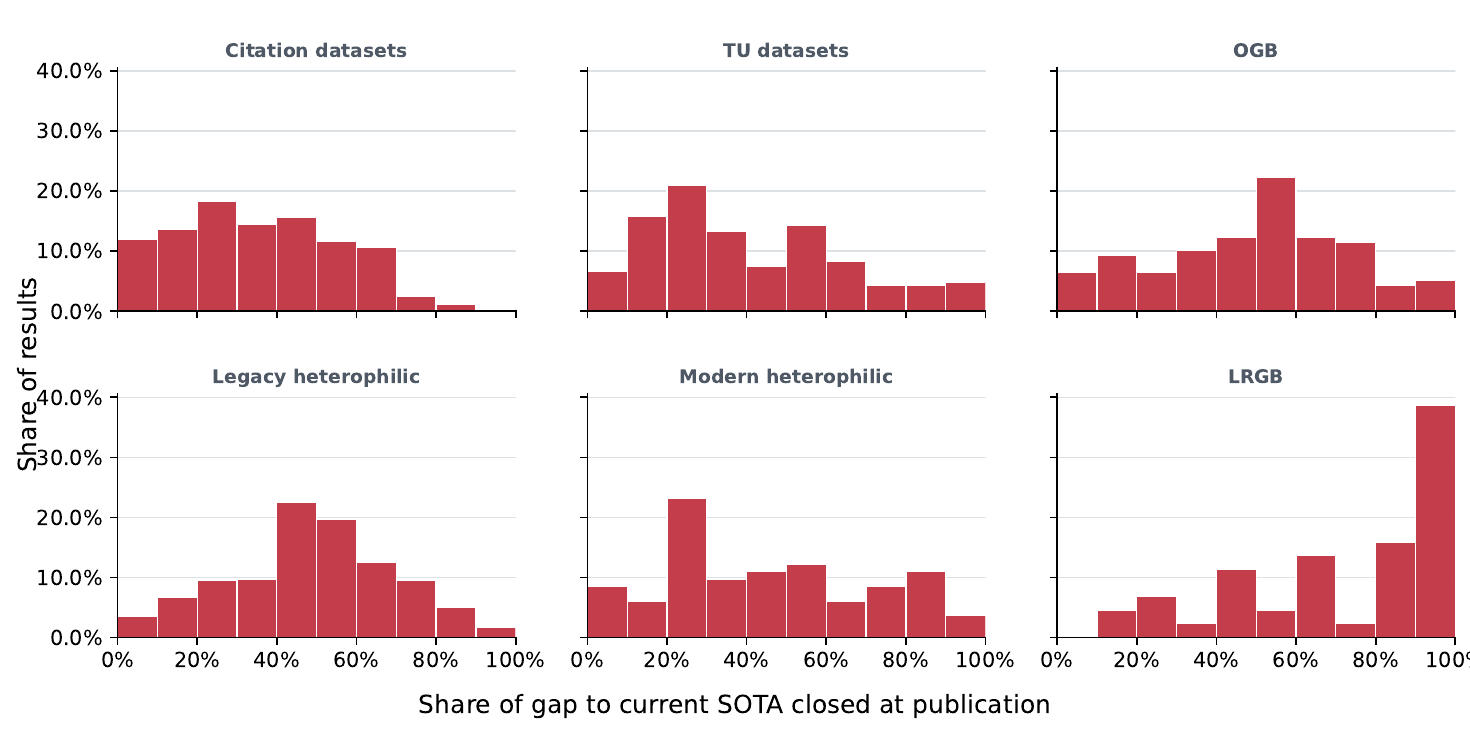}
    \caption{Relative reduction in a method’s distance to the SOTA using its corpus value instead of its paper-reported value, by dataset group.}
    \label{fig:gap_closed_appendix}
\end{figure}

We additionally present a detailed version of \cref{fig:gap_closed} split by dataset group as given in \cref{fig:stale_baselines,fig:stale_baselines_appendix}.
The most obvious feature is the peak at 100\% for LRGB, which is due to a simple GCN being SOTA on Peptides-struct \cite{whereDidTheGapGo} while being underreported by the original paper \cite{Dwivedi:2022lrgb}.
The plots in \cref{fig:gap_closed_appendix} show that the share of selected stale results closing at least half of the gap ranges from 26.2\% on citation datasets to 75.0\% on LRGB.
Compared to the figure in the main paper, the interpretation is more nuanced: while on LRGB those massive improvements are common, on other datasets the difference is more moderate but still substantial.
\citet{whereDidTheGapGo} showed that through underreported numbers on LRGB, the gains of transformers looked larger than they actually were.
Here, in \cref{fig:gap_closed_appendix} and \cref{fig:stale_baselines_appendix}, we showed that such underreported results are common throughout the corpus and that the underreporting is often substantial.

\subsection{Standard-Deviation Reporting}

\begin{figure}[t]
    \centering
    \includegraphics[width=.7\linewidth]{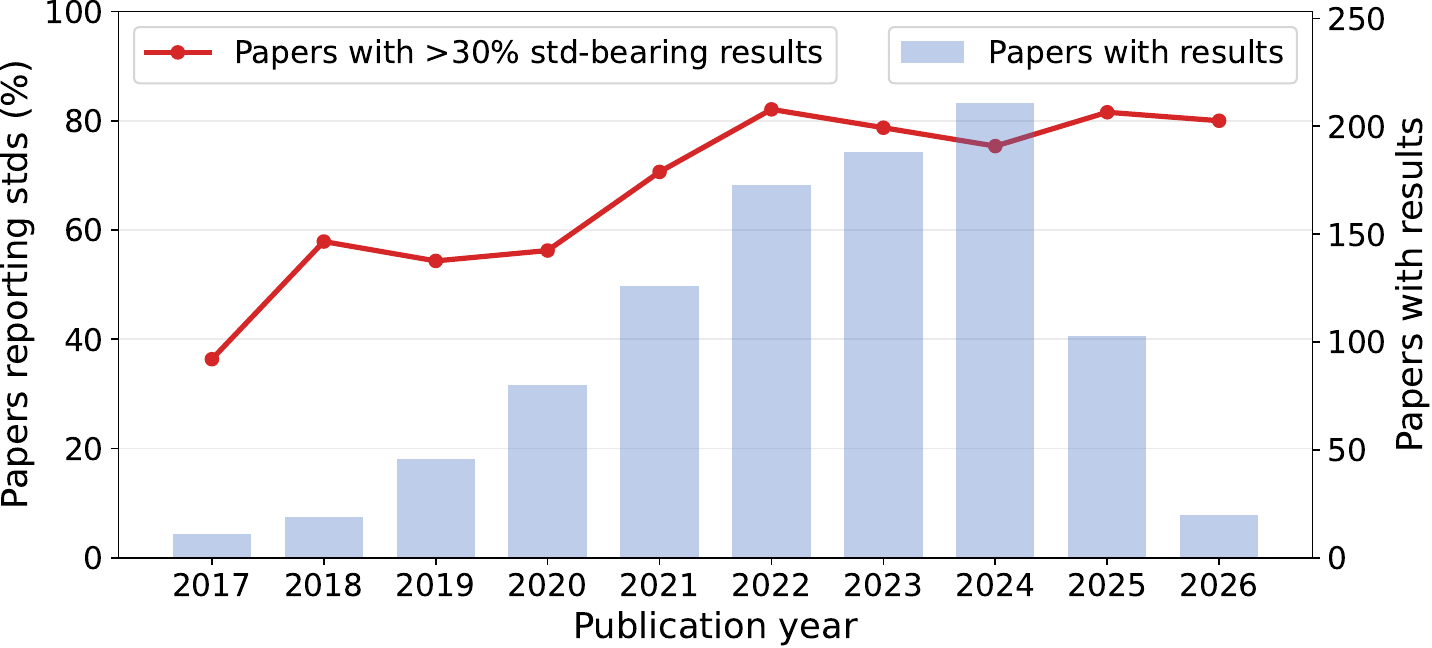}
    \caption{Fraction of papers reporting standard deviations. Starting with less than 40\% in 2017 it rose to $\sim\!80\%$ in 2022 and stagnated there since.}
    \label{fig:std-reporting}
\end{figure}

Based on the extracted data, we analyzed how many papers report standard deviations, counting papers with more than 30\% of results containing one.
Standard-deviation reporting improved from 36.4\% of papers in 2017 to roughly 80\% since 2022 (\cref{fig:std-reporting}).
Although standard deviations are now common in graph machine learning, around one in five papers still does not report them for its experiments.

\section{LLM Statement}

Language models were used during the development of the extraction pipeline and website, as well as in generating the plots. LLMs used include the GPT family as well as Composer 2.5 and Kimi-2.6.
GPT 5.6-sol was used during the writing process (grammar, proofreading, drafting an early version) and GPT 6 for polishing the camera ready version.

\end{document}